\documentclass[10pt,twocolumn,letterpaper]{article}

\usepackage{cvpr}              

\usepackage{amsthm}
\newtheorem{theorem}{Theorem}
\usepackage{pifont}
\usepackage{graphicx}
\usepackage{multirow}
\usepackage{tabularx}
    \newcolumntype{L}{>{\raggedright\arraybackslash}X}
\usepackage{float}

\usepackage{ctable}

\usepackage{pifont}

\usepackage{amsthm}

\definecolor{cvprblue}{rgb}{0.21,0.49,0.74}
\usepackage[pagebackref,breaklinks,colorlinks,allcolors=cvprblue]{hyperref}
\makeatletter
\usepackage{amsmath} %
\usepackage{graphicx} %
\usepackage[misc]{ifsym}
\def\paperID{4329} 
\def\confName{CVPR}
\def\confYear{2025}

\newcommand{\authornote}[1]{{%
  \let\thempfn\relax
  \footnotetext[0]{#1}
}}

\title{AutoLUT: LUT-Based Image Super-Resolution with Automatic Sampling and Adaptive Residual Learning}

\author{
Yuheng Xu$^*$ \quad Shijie Yang$^*$ \quad Xin Liu \quad Jie Liu\textsuperscript{\Letter} \quad Jie Tang \quad Gangshan Wu \\ 
State Key Laboratory for Novel Software Technology, Nanjing University, Nanjing 210023, China \\
{\tt\small superkenvery@gmail.com \quad \{sjyang,xinliu2023\}@smail.nju.edu.cn \quad
{\tt\small 
\{liujie,tangjie,gswu\}@nju.edu.cn}}
}

\begin{document}
\maketitle
\authornote{*: These authors contributed equally to this work.}
\authornote{\Letter: Corresponding author (liujie@nju.edu.cn).}
\begin{abstract}
In recent years, the increasing popularity of Hi-DPI screens has driven a rising demand for high-resolution images. However, the limited computational power of edge devices poses a challenge in deploying complex super-resolution neural networks, highlighting the need for efficient methods. While prior works have made significant progress, they have not fully exploited pixel-level information. Moreover, their reliance on fixed sampling patterns limits both accuracy and the ability to capture fine details in low-resolution images. To address these challenges, we introduce two plug-and-play modules designed to capture and leverage pixel information effectively in Look-Up Table (LUT) based super-resolution networks. Our method introduces Automatic Sampling (AutoSample), a flexible LUT sampling approach where sampling weights are automatically learned during training to adapt to pixel variations and expand the receptive field without added inference cost. We also incorporate Adaptive Residual Learning (AdaRL) to enhance inter-layer connections, enabling detailed information flow and improving the network’s ability to reconstruct fine details. Our method achieves significant performance improvements on both MuLUT and SPF-LUT while maintaining similar storage sizes. Specifically, for MuLUT, we achieve a PSNR improvement of approximately +0.20 dB improvement on average across five datasets. For SPF-LUT, with more than a 50\% reduction in storage space and about a 2/3 reduction in inference time, our method still maintains performance comparable to the original. Code is available at \href{https://github.com/SuperKenVery/AutoLUT}{https://github.com/SuperKenVery/AutoLUT}.
\end{abstract}    
\section{Introduction}
\label{sec:intro}
\begin{figure}[ht]
    \centering
\includegraphics[width=0.8\columnwidth,page=1]{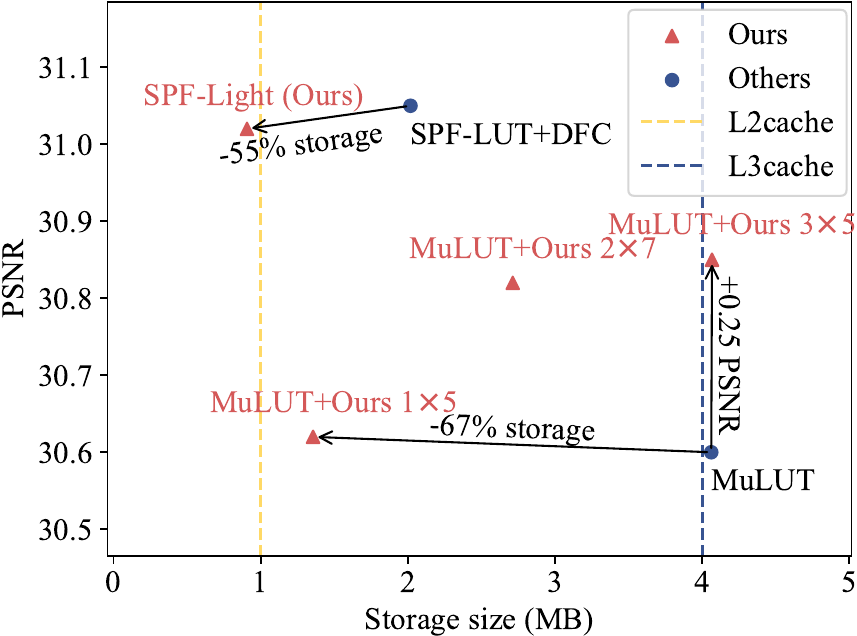}
    \caption{Performance-storage trade-offs for ×4 super-resolution
on Set5~\cite{DBLP:conf/bmvc/BevilacquaRGA12} compared with MuLUT~\cite{Li_2022_MuLUT} and SPF-LUT+DFC~\cite{Li_2024_CVPR}. As shown, our method can achieve the same level of performance at much lower storage requirement or much better performance with the same storage.}
    \label{fig:perf-vs-storage}
\end{figure}
Image Super-Resolution (SR) is the process of enhancing the resolution of an image by reconstructing a High-Resolution (HR) version from Low-Resolution (LR) inputs. SR methods based on Deep Neural Networks (DNNs) have demonstrated remarkable success \cite{DBLP:conf/eccv/DongLHT14,DBLP:conf/cvpr/KimLL16a,DBLP:conf/eccv/DongLT16,DBLP:conf/cvpr/LimSKNL17,DBLP:conf/cvpr/ZhangTKZ018,DBLP:conf/eccv/ZhangLLWZF18,DBLP:conf/eccv/WangYWGLDQL18}, but these deep learning techniques often entail substantial computational demands, making them difficult to deploy.


A significant contribution to addressing these challenges is the SR-LUT \cite{DBLP:conf/cvpr/JoK21}, which leverages Look-Up Tables (LUTs) to expedite neural network inference. SR-LUT trains a convolutional SR network, then caches its results into an LUT, looks them up during inference, thus reducing computational burden. However, a major limitation of this method is that the size of LUT grows exponentially with the number of input pixels, which limits the receptive field and degrades performance. To overcome this limitation, the subsequent MuLUT \cite{Li_2022_MuLUT} uses multiple LUTs, each processing four input pixels sampled from different locations within the image, forming a $3 \times 3$ receptive field. MuLUT further integrates different strategies within a single LUT group, where each group initially has a receptive field of $3 \times 3$. By incorporating a rotation ensemble strategy, this receptive field is expanded to $5 \times 5$. Additionally, MuLUT introduces a two-stage SR network that consists of two LUT groups, resulting in a final receptive field of $9 \times 9$. To achieve better performance, SPF-LUT \cite{Li_2024_CVPR} further enlarges the network scale and then proposes a diagonal-first compression technique that preserves the most important parts of the LUT while compressing less important parts more aggressively. This reduces LUT size by approximately 90\%.
However, SPF-LUT expands the receptive field by deepening the network rather than altering the sampling strategy, it retains the manually designed sampling patterns used by MuLUT, which may not align well with the natural variations in data, limiting the model’s ability to capture fine details accurately.  Additionally, previous models did not incorporate residual layers, as adding skip connections would push the LUT values beyond their valid range, making it difficult to preserve their integrity. This lack of residual layers further restricts the flow of information between layers, limiting the model’s ability to learn complex features.

To address these issues, we propose two novel plug-and-play modules for LUT-based SR networks: Automatic Sampling (AutoSample) and Adaptive Residual Learning (AdaRL). AutoSample automatically learns how to sample during training, transforming static pixel extractions into learnable pixel abstractions. This approach allows the receptive field to automatically adapt based on the input during training, providing greater flexibility. AdaRL introduces a modified residual layer that enhances the fusion of information between network layers, improving the flow of detailed information without exceeding the storage constraints of LUTs. Experiments show that our method achieves better performance without additional consumption. As shown in Fig.~\ref{fig:perf-vs-storage}, with more than a 50\% reduction in storage space and a 2/3 reduction in inference time, our method still maintains comparable performance. Furthermore, its plug-and-play design allows for easy integration into various networks, providing a performance boost across the entire family of LUT-based SR networks.

The contributions of this work are as follows:

1) We propose Automatic Sampling (AutoSample), enabling learnable pixel abstractions that allow for a larger receptive field.

2) We introduce the AdaRL layer, adapting the residual technique to LUT-based super-resolution networks to improve the flow of information between layers and enhance model performance.

3) We design both the AutoSample and AdaRL layer as plug-and-play modules, enabling flexible integration into LUT-based networks.

 4) Quantitative and qualitative results demonstrate that our method achieves improved performance while significantly reducing storage space and inference time, making it highly suitable for deployment on resource-limited devices.

\section{Related Works}
\begin{figure*}
    \centering
    \includegraphics[width=1\linewidth,page=1]{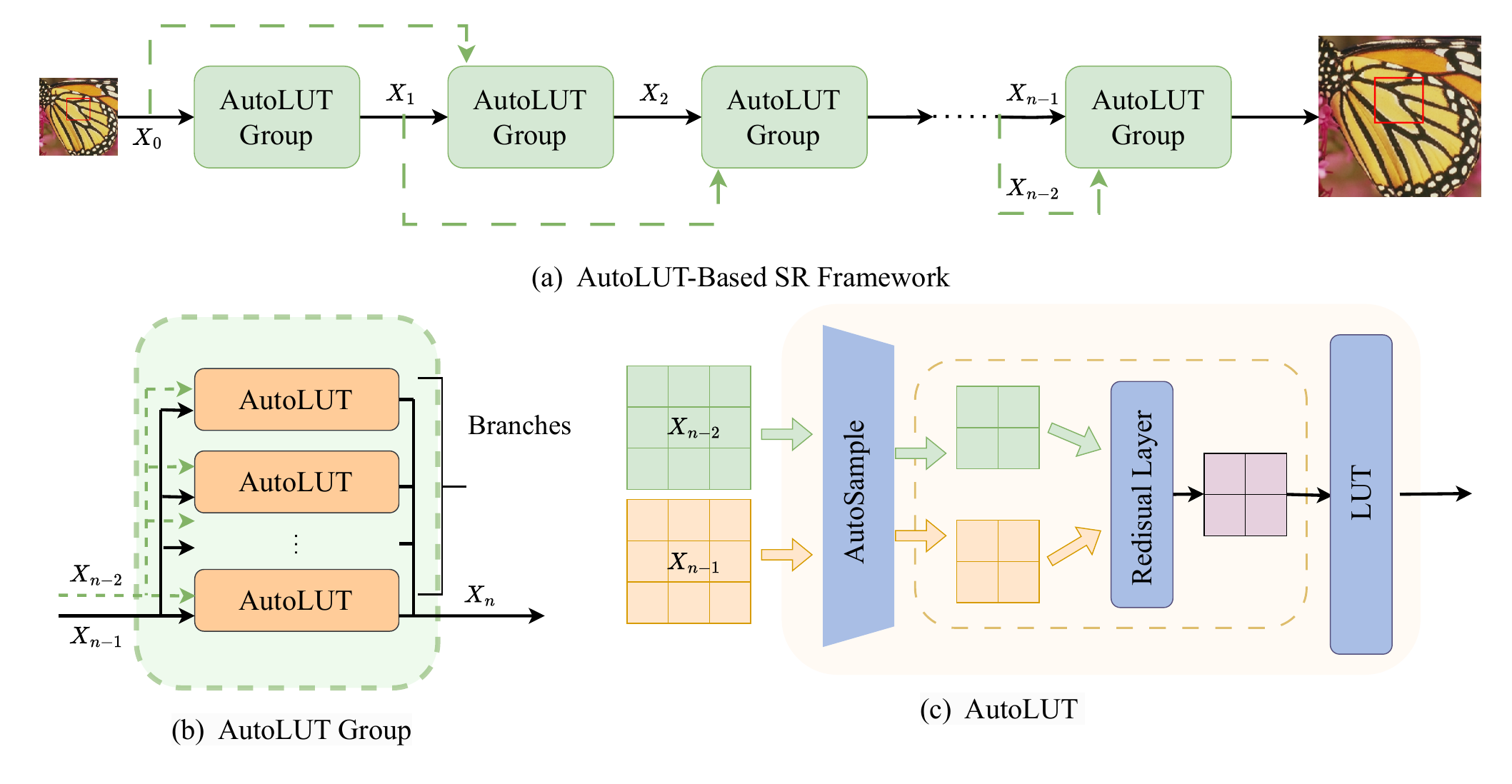}
    \caption{Overview of the AutoLUT-based SR framework. The figure illustrates the process of enhancing LUT-based super-resolution models by integrating the AutoLUT. (a) The AutoLUT-based SR framework replaces traditional LUT Group with AutoLUT Group for improved flexibility. (b) The AutoLUT Group can flexibly adjust the number of branches according to the requirements, enabling more efficient processing of diverse information. (c) The AutoLUT, where the inputs $X_{n-1}$ and $X_{n-2}$ pass through the AutoSample, followed by a combination using learnable residual weights. The final output is processed through the basic LUT.}
    \label{fig:framework}
\end{figure*}
\textbf{Classical SR Methods. } Classical super-resolution algorithms, such as nearest interpolation, bilinear interpolation, and bicubic interpolation \cite{Keys1981CubicCI}, are widely used in emulators and monitors due to their low computational cost, making them suitable for resource-constrained environments. However, they often produce visible artifacts like blurring and aliasing, especially compared to modern DNN-based methods. Additionally, there are exemplar-based methods \cite{DBLP:journals/cga/FreemanJP02,DBLP:journals/ijcv/FreemanPC00,DBLP:journals/jvcir/Qiu00,DBLP:journals/tip/XiongSW10a,DBLP:journals/tmm/XiongXSW13} and sparse-coding methods \cite{DBLP:journals/tip/YangWHM10,DBLP:conf/cas/ZeydeEP10,DBLP:conf/iccv/TimofteDG13,DBLP:conf/accv/TimofteSG14,DBLP:journals/tci/RomanoIM17}. While sparse-coding methods can perform well, they are computationally intensive. 
Other techniques, such as gradient field sharpening \cite{DBLP:journals/tip/SongXLXWG18} and displacement fields\cite{DBLP:journals/tip/WangWP14}, also show promise but face challenges with image clarity and high computational requirements.

\textbf{DNN Based SR Methods. } With the development of deep learning, many DNN based image super-resolution algorithms have achieved significant restoration performance \cite{DBLP:conf/eccv/AhnKS18, DBLP:conf/cvpr/ChenXTZW19, DBLP:conf/cvpr/ChengXC0Z21, DBLP:conf/cvpr/KimLL16a, DBLP:conf/cvpr/XiaoFHCX21, DBLP:conf/mm/XiaoXFLZ20, DBLP:conf/mm/XuXYZX21, DBLP:conf/iccv/ZhangLX19, DBLP:conf/eccv/ZhangLLWZF18, DBLP:journals/pami/DongLHT16, DBLP:conf/cvpr/LimSKNL17, DBLP:conf/eccv/WangYWGLDQL18, DBLP:conf/nips/HoJA20}
. Early models like SRCNN \cite{DBLP:journals/pami/DongLHT16}, introduced simple yet effective architectures. The development of residual networks (ResNets) \cite{DBLP:conf/cvpr/HeZRS16} enabled the training of deeper networks by alleviating gradient issues, leading to improvements in SR tasks. EDSR \cite{DBLP:conf/cvpr/LimSKNL17} and ESRGAN \cite{DBLP:conf/eccv/WangYWGLDQL18} applied these techniques, achieving state-of-the-art performance by incorporating residual learning and GANs for more realistic outputs. More recently, diffusion-based models \cite{DBLP:conf/nips/HoJA20} have introduced promising generative capabilities to SR. However, while DNN-based methods demonstrate strong restoration abilities, they require significant computational resources due to their complex architectures with large parameter counts.

\textbf{Look-Up Table Based SR Methods. } Recently, the use of Look-Up Tables (LUTs) has garnered increasing attention in the field of image restoration due to their low computational cost during inference. SR-LUT \cite{DBLP:conf/cvpr/JoK21}  pioneered LUT-based image super-resolution, laying the groundwork for this approach. Building upon this foundation, SPLUT \cite{DBLP:conf/eccv/MaZZL22} improved performance by splitting images into Most Dignificant Bits (MSB) and Least Significant Bits (LSB) and processing each part separately before recombining. MuLUT \cite{Li_2022_MuLUT} further extended the receptive field by utilizing multiple LUTs, complementing different sampling methods to achieve a larger receptive field and higher complexity. Most recently, SPF-LUT \cite{Li_2024_CVPR} introduced a diagonal-first compression technique that reduces redundancy by leveraging the commonality of neighboring pixels. However, since SPF-LUT uses the same fixed sampling strategy as MuLUT, the receptive field of a single LUT group remains confined to $3 \times 3$, limiting the flexibility for further improvements in feature aggregation.


\section{Method}

\subsection{Overview}
To enhance the performance of LUT-based super-resolution models, several architectural improvements have been proposed. SR-LUT \cite{DBLP:conf/cvpr/JoK21} uses a single LUT for mapping inputs to outputs. MuLUT \cite{Li_2022_MuLUT} extends this by using two LUT groups, each with three different sampling strategies, forming multiple branches to enhance the model’s flexibility. SPF-LUT \cite{Li_2024_CVPR} takes this a step further by incorporating even more LUT groups, expanding the model's depth and increasing its capacity to handle complex tasks.

Building upon these advancements, we propose the AutoLUT-based SR framework, a novel architecture that replaces traditional LUT Group with the proposed AutoLUT Group. As shown in Fig.~\ref{fig:framework}
, the input to the $n^{th}$ group consists of $X_{n-1}$ and $X_{n-2}$, facilitating better communication of fine-grained information across different layers.
Previous methods rely on a manually designed, fixed three-branch configuration per LUT group, with each branch selecting four fixed pixels to cover a $3 \times 3$ receptive field, requiring exactly three branches for full coverage. This hand-crafted approach can not adjust to data variations. In contrast, our AutoLUT automatically learns the optimal sampling strategy directly from the network, allowing a single branch to cover a $k \times k$ receptive field and enabling flexible adjustment of branch numbers based on requirements. Specifically, we pass $X_{n-1}$ and $X_{n-2}$ through the AutoSample layer, which extracts feature information across the entire receptive field, adapting as needed to achieve full coverage. These are then combined using the AdaRL, and the final output is processed through the basic LUT. This approach enables AutoLUT to create adaptable, learnable representations of pixel values, adjusting branch configurations based on input requirements.

\begin{figure*}
    \centering
    \includegraphics[width=1\linewidth,page=1]{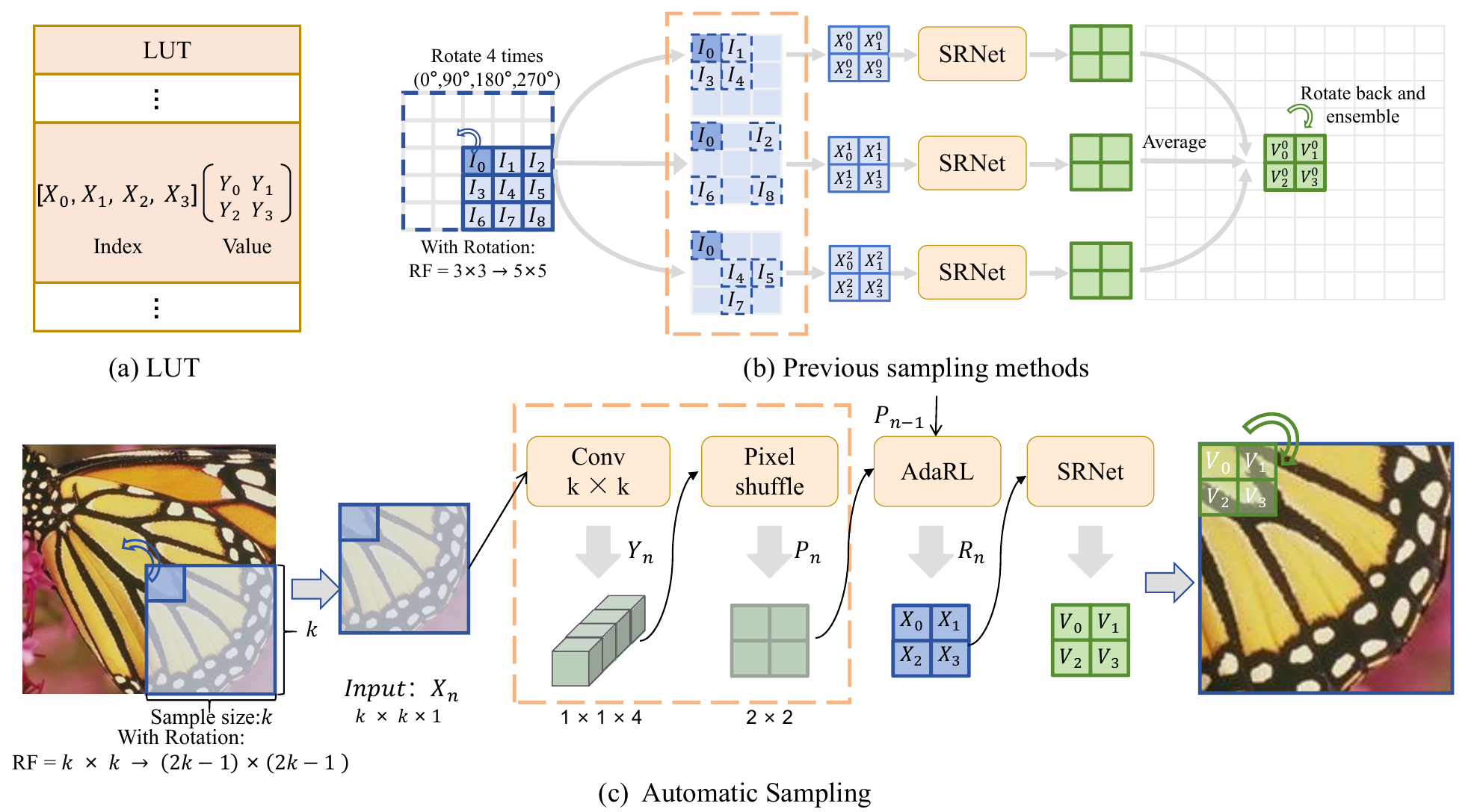}
    \caption{Comparison of sampling strategies on $\times$2 super-resolution. (a) A LUT is a data structure composed of index-value pairs, the sampled indices are used to retrieve their corresponding values. (b) The sampling methods in MuLUT \cite{Li_2022_MuLUT} and SPF-LUT \cite{Li_2024_CVPR}, use three distinct sampling strategies, each selecting four fixed pixels from a $3 \times 3$ input window. Together, these three sampling configurations fully cover the $3 \times 3$ window. After applying a rotation ensemble, where the results from four rotated versions are averaged, the receptive field expands from $3 \times 3$ to $5 \times 5$. The pixel \( I_0 \) in the low-resolution image corresponds to pixel \([ V_{0}^0, V_{0}^1, V_{0}^2, V_{0}^3 ] \) in the image after super-resolution.
(c) The learnable AutoSample strategy selects pixels by learning sampling weights through convolution during training,  using pixel shuffle to expand the receptive field to $(2k-1) \times (2k-1)$. Finally, the desired index is obtained by applying AdaRL.}
    \label{fig:autosample}
\end{figure*}

\subsection{Automatic Sampling (AutoSample)}
As shown in Fig.~\ref{fig:autosample}(a), a LUT is a data structure composed of index-value pairs, where sampled indices are used to retrieve their corresponding values. The sampling strategy, which determines how indices are selected, plays a crucial role in mapping them effectively to their values. This directly influences the quality of feature extraction by controlling which information is captured from the input data.

Current methods typically select four fixed pixels directly from the original input, mapping these pixels to their corresponding values, which limits flexibility in handling diverse input. As shown in Fig.~\ref{fig:autosample}(b), MuLUT \cite{Li_2022_MuLUT} and SPF-LUT \cite{Li_2024_CVPR} employ three distinct sampling strategies, each of which selects four specific pixels from a $3 \times 3$ input window. These three sampling configurations collectively cover the entire $3 \times 3$ region, and with the application of a rotation ensemble~\cite{DBLP:conf/cvpr/JoK21}, the network effectively expands its receptive field from $3 \times 3$ to $5 \times 5$. However, because these strategies are predefined and fixed, they lack the flexibility to adapt to diverse image content or feature distributions. This rigidity restricts the models from adjusting their receptive fields based on specific input characteristics, which may lead to suboptimal performance.

To address these limitations, our approach introduces a learnable sampling strategy, where the network automatically learns how to sample from the input during training. The specific structure is shown in Fig.~\ref{fig:autosample}(c). In our framework, we define the sample size as \( k \), and the input \( X_n \) with dimensions \( k \times k \times 1 \) undergoes a convolution operation:
\begin{equation}
    Y_n = \text{Conv}(X_n, W),
    \label{conv}
\end{equation} 
where \( Y_n \in \mathbb{R}^{1 \times 1 \times 4}\). In the training process, the sampling weights \( W \) with a shape of \( k \times k \times 1 \times 4 \) are learned, enabling each output channel in \( Y_n \) to represent a weighted combination of the \( k \times k \) spatial region of the input \( X_n \). Specifically, for each output channel \( c \), the value is computed as a weighted sum of the corresponding input pixels:

\begin{equation}
    Y_n^{(c)} = \sum_{i=1}^{k} \sum_{j=1}^{k} X_n^{(i,j)} \cdot W^{(i,j,1,c)}.
\end{equation}
 To ensure that the weighted sum does not exceed the range of [0, 255], we apply a softmax operation across the \( k \times k \) spatial dimensions of \( W \) before the convolution:

\begin{equation}
W^{(i,j,1,c)} = \frac{\exp(W^{(i,j,1,c)})}{\sum_{i'=1}^{k} \sum_{j'=1}^{k} \exp(W^{(i',j',1,c)})}.
\end{equation}
This normalization is performed over the spatial dimensions \( (0, 1) \), which ensures that the weights for each output channel in \( W \) are non-negative and sum to one, ensuring that the convolution results stay within [0, 255]. 

To justify the effectiveness of our method, we first introduce convex combination theory \cite{bertsekas2003convex} as follows:
\begin{theorem} 
\label{convex}
Given \( n \) values \( a_1, a_2, \dots, a_n \) such that each \( a_i \) lies within an interval \( [a, b] \), and corresponding non-negative weights \( \lambda_1, \lambda_2, \dots, \lambda_n \) that sum to 1 (i.e., \( \sum_{i=1}^{n} \lambda_i = 1 \), \( \lambda_i \geq 0 \)), the weighted sum of these values is: $\tilde{a} = \sum_{i=1}^{n} \lambda_i a_i$. 
It will also lie within the interval \( [a, b] \), i.e.,
\[ \min(a_1, a_2, \dots, a_n) \leq \tilde{a} \leq \max(a_1, a_2, \dots, a_n).\] \end{theorem}
Based on this, our method ensures that the weighted combination of input pixels remains within the [0, 255].

Following the convolution, the \( Y_n \) output undergoes a pixel shuffle operation to produce a \( 2 \times 2 \) grid of pixels, represented as \( P_n \):
\begin{equation}
 P_n = \text{Pixelshuffle}(Y_n).
\end{equation}
where \( P_n \in \mathbb{R}^{2 \times 2} \). This operation upscales \( Y_n \) by a factor of 2, redistributing the four output channels in \( Y_n \) to form a \( 2 \times 2 \) spatial arrangement in \( P_n \). Each of the four pixels in \( P_n \) is thus a weighted combination of the \( k^2 \) surrounding pixels from the original input.
At this point, the receptive field is \( k \times k \), the effective receptive field size expands to \( (2k - 1) \times (2k - 1) \) after a rotation ensemble strategy \cite{DBLP:conf/cvpr/JoK21}. 

By replacing the fixed sampling patterns with AutoSample, our approach provides greater flexibility in both the network's architecture and its configuration. Unlike previous methods that rely on three distinct, fixed sampling strategies to cover the entire receptive field, our method can use a single AutoSample branch to achieve this. Moreover, our approach allows for the flexible adjustment of the sample size, which directly influences the receptive field. This flexibility not only simplifies the sampling process but also enables the network to automatically learn the sampling strategy. By adding more branches, the network can learn more sampling methods, enhancing its ability to capture diverse features from different input scenarios.

\subsection{Adaptive Residual Learning (AdaRL)}
In LUT-based networks, directly using residual connections 
may cause the combination of two values within the [0, 255] range to exceed the LUT's input range, leading to a dramatic increase in the LUT size. While normalizing the residuals or directly clamping them can prevent the LUT from becoming too large, it results in a loss of precision, which may negatively impact performance~\cite{DBLP:conf/cvpr/LimSKNL17}.

To address this, we propose Adaptive Residual Learning (AdaRL), which performs a weighted average between the current and previous pixel values, maintaining values within a manageable range and preventing the LUT from expanding excessively. AdaRL learns spatially varying weights for each pixel, denoted as \( W_{Residual} \) with a shape 2 $\times$ 2, which determine the contribution of the current and previous pixel values. The weights are clamped between 0 and 1 to ensure a valid range. As shown in Fig.~\ref{fig:framework}(c),
the inputs \( X_{n-1} \) and \( X_{n-2} \) undergo the AutoSample process, which extracts relevant pixel information from the input patch:
\begin{equation}
    P_{n-2} = \text{AutoSample}(X_{n-2}),
\end{equation}
\begin{equation}
    P_{n-1} = \text{AutoSample}(X_{n-1}).
\end{equation}
Then, the sampled outputs \( P_{n-1} \) and \( P_{n-2} \), each of size \( 2 \times 2 \), are combined using the learnable residual weights \( W_{\text{Residual}} \) as follows:
\begin{equation}
    R_{n-1}^{(i,j)} = (1 - W_{\text{Residual}}^{(i,j)}) \odot P_{n-1}^{(i,j)} + W_{\text{Residual}}^{(i,j)} \odot P_{n-2}^{(i,j)},
    \label{residual_conbine}
\end{equation} 
which can be viewed as a convex combination.

According to the Theorem~\ref{convex}, \( P_{n-1} \) and \( P_{n-2} \) are constrained within the range \([0, 255]\), and \( W_{Residual} \in [0, 1] \), the output \( R_{n-1} \) will lie within the range:
\begin{equation}
R_{n-1} \in [\min(P_{n-1}, P_{n-2}), \max(P_{n-1}, P_{n-2})],
\end{equation}
ensuring that \( R_{n-1} \) remains within the range \([0, 255]\).

This confirms that AdaRL preserves precise residual features without causing any issues with LUT size. Therefore, the adaptive weighting in AdaRL ensures that the output \( R_{n-1} \) stays within the desired range of [0, 255], maintaining the quality of the image reconstruction and preventing LUT size expansion.
\section{Experiments and Results}
\begin{table*}[htb]
    \centering    \caption{Quantitative comparison of PSNR/SSIM and storage size on standard benchmark datasets for ×4 super-resolution. We applied AutoSample and AdaRL to SPF-LUT \cite{Li_2024_CVPR} and MuLUT \cite{Li_2022_MuLUT}. Using our method on MuLUT with 3 branches and sample size 5, we achieved better performance while maintaining nearly the same storage size. On SPF-LUT, our method achieved comparable performance with less than half the storage. \colorbox{gray!30}{The gray background} highlights our methods. For MuLUT and SPF-LUT, the best results are in \textbf{bold}.}
    \def\arraystretch{1.1}
\resizebox{\textwidth}{!}{  
    \begin{tabular}{cccccccc}
        \hline
         & \textbf{Method} & \textbf{Storage Size} & \textbf{Set5} \cite{DBLP:conf/bmvc/BevilacquaRGA12} & \textbf{Set14} \cite{zeyde2010single} & \textbf{BSDS100} \cite{DBLP:conf/iccv/MartinFTM01} & \textbf{Urban100} \cite{DBLP:conf/cvpr/HuangSA15} & \textbf{Manga109} \cite{DBLP:journals/mta/MatsuiIAFOYA17} \\
         \hline
         \multirow{7}{*}{Classical}  
         & Nearest & - & 26.25/0.7372 & 24.65/0.6529 & 25.03/0.6293 & 22.17/0.6154 & 23.45/0.7414 \\
         & Zeyde et al. \cite{DBLP:conf/cas/ZeydeEP10} & - & 26.69/0.8429 & 26.90/0.7354 & 26.53/0.6968 & 23.90/0.6962 & 26.24/0.8241 \\
         & Bilinear & - & 27.55/0.7884 & 25.42/0.6792 & 25.54/0.6460 & 22.69/0.6346 & 24.21/0.7666 \\
         & Bicubic & - & 28.42/0.8101 & 26.00/0.7023 & 25.96/0.6672 & 23.14/0.6574 & 24.91/0.7871 \\
         & NE + LLE \cite{DBLP:conf/cvpr/ChangYX04} & 1.434MB & 29.62/0.8404 & 26.82/0.7346 & 26.49/0.6970 & 23.84/0.6942 & 26.10/0.8195 \\
         & ANR \cite{DBLP:conf/iccv/TimofteDG13} & 1.434MB & 29.70/0.8422 & 26.86/0.7386 & 26.52/0.6992 & 23.89/0.6964 & 26.18/0.8214 \\
         & A+ \cite{DBLP:conf/accv/TimofteSG14} & 15.17MB & 30.27/0.8602 & 27.30/0.7498 & 26.73/0.7088 & 24.33/0.7189 & 26.91/0.8480 \\
         \hline
         \multirow{3}{*}{MuLUT}
         & MuLUT \cite{Li_2022_MuLUT} & 4.062MB & 30.60/0.8653 & 27.60/0.7541 & 26.86/0.7110 & 24.46/0.7194 & 27.90/0.8633 \\
         & MuLUT+DFC \cite{Li_2024_CVPR} & \textbf{0.407MB} & 30.55/0.8642 & 27.56/0.7532 & 26.83/0.7104 & 24.41/0.7177 & 27.82/0.8613 \\
         &  \cellcolor{gray!30}MuLUT+ours & \cellcolor{gray!30}4.067M & \cellcolor{gray!30}\textbf{30.85/0.8699} & \cellcolor{gray!30}\textbf{27.77/0.7584} & \cellcolor{gray!30}\textbf{26.96/0.7144} & \cellcolor{gray!30}\textbf{24.60/0.7257} & \cellcolor{gray!30}\textbf{28.27/0.8706}\\
         \hline
         \multirow{3}{*}{SPF-LUT}
         & SPF-LUT \cite{Li_2024_CVPR} & 17.284MB & \textbf{31.11/0.8764} & \textbf{27.92/0.7640} & \textbf{27.10/0.7197} & \textbf{24.87/0.7378} & \textbf{28.68/0.8796} \\
         & SPF-LUT+DFC \cite{Li_2024_CVPR} & 2.018MB & 31.05/0.8755 & 27.88/0.7632 & 27.08/0.7190 & 24.81/0.7357 & 28.58/0.8779 \\
         & \cellcolor{gray!30}SPF-Light & \cellcolor{gray!30}\textbf{0.907MB} & \cellcolor{gray!30}31.02/0.8751 & \cellcolor{gray!30}27.88/0.7629 & \cellcolor{gray!30}27.07/0.7186 & \cellcolor{gray!30}24.78/0.7342 & \cellcolor{gray!30}28.54/0.8769 \\
         \hline
         \multirow{4}{*}{DNN} 
         & RRDB \cite{DBLP:conf/eccv/WangYWGLDQL18} & 63.942MB & 32.68/0.8999 & 28.88/0.7891 & 27.82/0.7444 & 27.02/0.8146 & 31.57/0.9185 \\
         & EDSR \cite{DBLP:conf/cvpr/LimSKNL17} & 164.396MB & 32.46/0.8968 & 28.80/0.7876 & 27.71/0.7420 & 26.64/0.8033 & 31.02/0.9148 \\
         & RCAN \cite{DBLP:conf/eccv/ZhangLLWZF18} & 59.74MB & 32.61/0.8999 & 28.93/0.7894 & 27.80/0.7436 & 26.85/0.8089 & 31.45/0.9187 \\
         & SwinIR \cite{DBLP:conf/cvpr/ZhangTKZ018} &    170.4MB & 
         32.44/0.8976 & 28.77/0.7858 & 27.69/0.7406 & 26.47/0.7980 & 30.92/0.9151 \\
         \hline
    \end{tabular}
    }
    \label{table:sr-methods-perf}
\end{table*}
\subsection{Experiment settings}
\begin{figure}[t!]
    \centering
    \includegraphics[width=0.8\columnwidth,page=1]{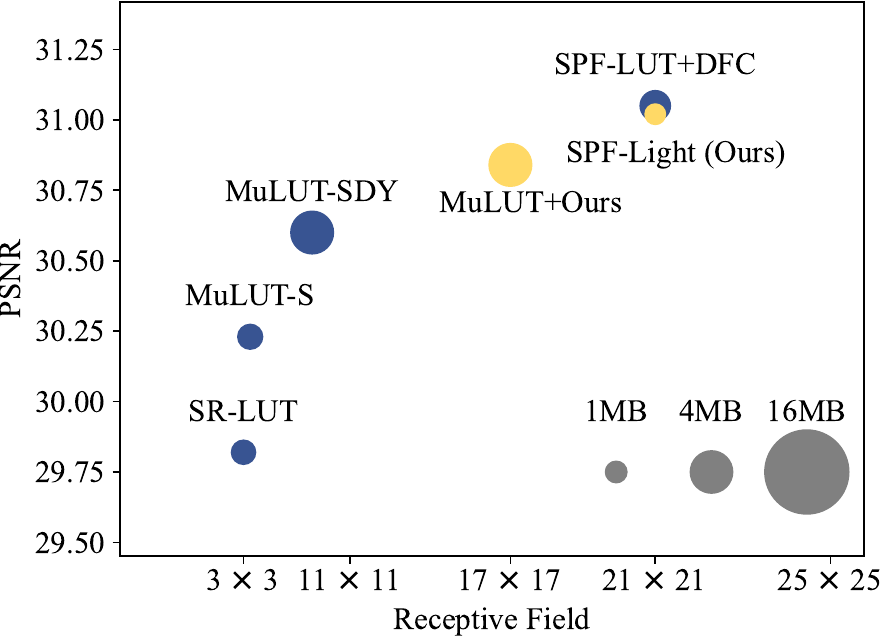}
    \caption{Receptive field and storage comparisons of LUTs. The size of the bubble represents the storage size.}
    \label{fig:lut-scaling-law}
\end{figure}
\textbf{Training Settings.} We trained our models on DIV2K~\cite{Agustsson_2017_CVPR_Workshops} dataset and we do evaluation on Set5~\cite{DBLP:conf/bmvc/BevilacquaRGA12}, Set14~\cite{zeyde2010single}, BSDS100~\cite{DBLP:conf/iccv/MartinFTM01}, Urban100~\cite{DBLP:conf/cvpr/HuangSA15} and Manga109~\cite{DBLP:journals/mta/MatsuiIAFOYA17}. For training, we used the Adam optimizer, chosen for its adaptive learning rate capabilities. The weight decay was set to 0, and the learning rate was initialized at $10^{-3}$. We randomly crop images to $ 48 \times 48 $ patches with a batch size of 32. After training, the models were exported as Look-Up Tables (LUTs). Where necessary, DFC \cite{Li_2024_CVPR} compression was applied to reduce the LUT size. Following this, a LUT-aware fine-tuning phase was performed \cite{Li_2022_MuLUT}, during which the LUT contents were treated as trainable parameters, allowing for further optimization of the network’s performance. During fine-tuning, the AutoSample and AdaRL were fine-tuned together.

To verify the effectiveness of our modules, we apply them to MuLUT \cite{Li_2022_MuLUT} and SPF-LUT \cite{Li_2024_CVPR}. Specifically, we conduct experiments with different parameter combinations, including the number of branches and the sample size
$k$. In our setup, ``MuLUT+Ours \(b \times k\)'' denotes the specific configuration with the number of branches \(b\) and sample size \(k\). Throughout the following sections, MuLUT+Ours specifically implies the configuration with 3 branches and sample size 5. Additionally, we implement a lightweight version of SPF-LUT, called SPF-Light, by replacing and reducing the number of branches in each group to one. This adjustment is intended to significantly reduce storage space requirements, allowing us to explore the efficiency of our method in optimizing model size without compromising performance.

\textbf{Comparison Models and Metrics.} We evaluate our approach against a variety of super-resolution methods, including interpolation-based techniques (nearest neighbor, bilinear, bicubic interpolation), sparse coding methods (NE+LLE \cite{DBLP:conf/cvpr/ChangYX04}, ANR \cite{DBLP:conf/iccv/TimofteDG13} and A+ \cite{DBLP:conf/accv/TimofteSG14}) and DNN methods (RRDB \cite{DBLP:conf/eccv/ZhangLLWZF18}, EDSR \cite{DBLP:conf/cvpr/LimSKNL17}, RCAN \cite{DBLP:conf/eccv/ZhangLLWZF18}, SwinIR \cite{DBLP:conf/cvpr/ZhangTKZ018} ). We additionally compare with MuLUT \cite{Li_2022_MuLUT} and SPF-LUT \cite{Li_2024_CVPR}. We test Peak Signal-to-Noise Ratio (PSNR) and Structural Similarity Index Measure (SSIM) for quantitative evaluation.
\begin{figure*}
    \centering
\includegraphics[width=0.9\linewidth,page=1]{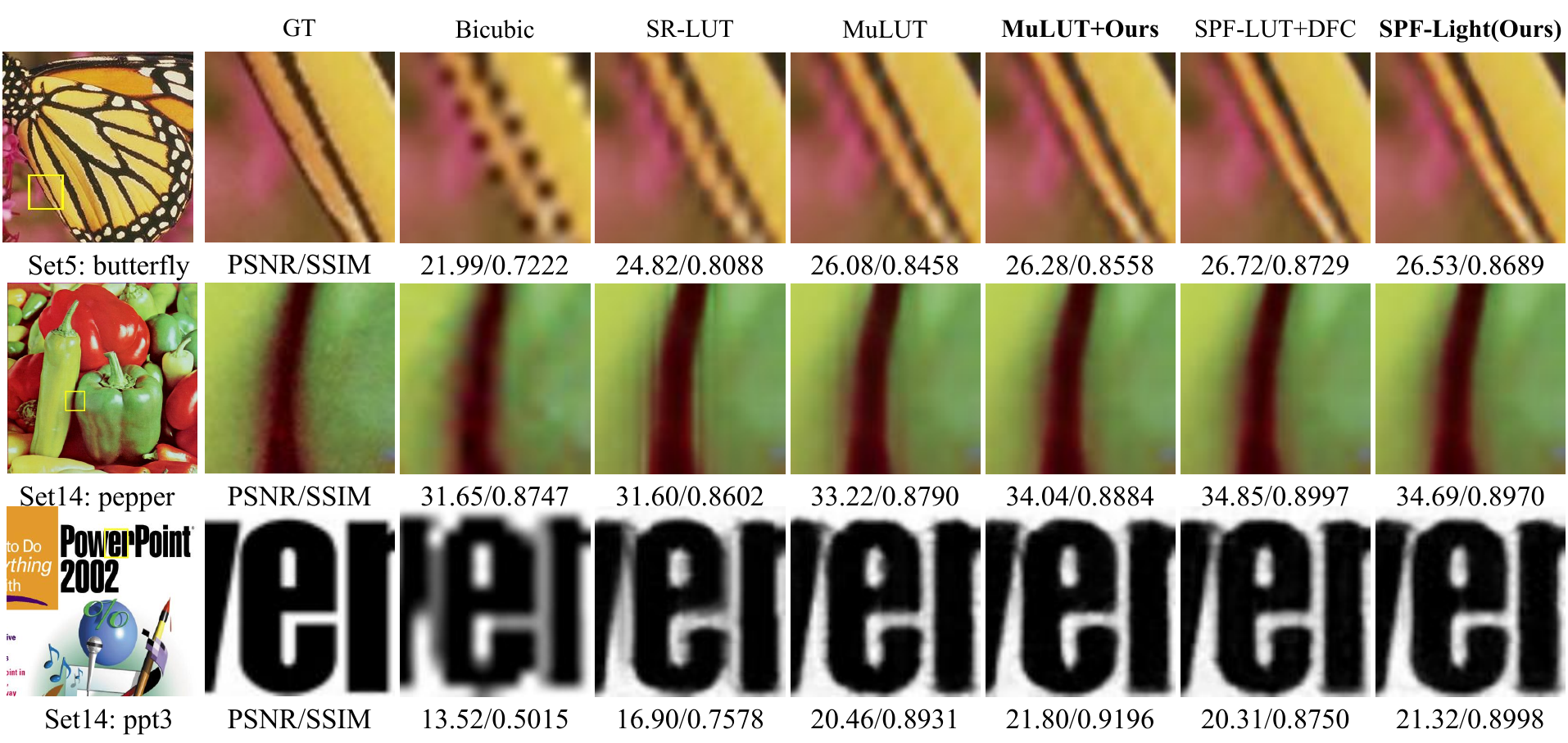}
    \caption{Qualitative comparison for $\times$4 super-resolution on benchmark datasets. Our methods are in \textbf{bold}. }
    \label{vis}
\end{figure*}
\subsection{Experimental Results and Analysis}
\textbf{Quantitative Comparison.}
We evaluate our framework on five datasets, using PSNR and SSIM as metrics. As shown in Table.~\ref{table:sr-methods-perf}, when our proposed modules are integrated into MuLUT, we observe a +0.20 dB improvement in PSNR on average across the datasets. Specifically, on the Set5 dataset, the PSNR increases by 0.25 dB,  on the Manga109 dataset, the PSNR improves by 0.37 dB. These results clearly demonstrate that our method provides substantial performance improvements, with a negligible increase in storage size (from 4.062 MB to 4.067 MB). We conducted a further visual analysis in Fig.~\ref{fig:perf-vs-storage} and Fig.~\ref{fig:lut-scaling-law}. As shown, SPF-Light reduces storage by 55\% while maintaining performance comparable to SPF-LUT+DFC (Fig.~\ref{fig:perf-vs-storage}), with the same receptive field (Fig.~\ref{fig:lut-scaling-law}). Additionally, MuLUT+Ours greatly expands the receptive field without increasing storage, resulting in a 0.25 dB PSNR improvement on Set5. Moreover, MuLUT+Ours 1$\times$5 reduces storage by about 67\% compared to MuLUT while exceeding its PSNR performance. Our method maintains the performance while significantly reducing storage, offering a more efficient solution for LUT-based super-resolution networks.

\begin{table*}[h!]
    \centering
    \caption{Ablation study on different branch numbers and sample sizes for $\times$4 super-resolution on MuLUT.}
    \def\arraystretch{1.1}
    \resizebox{\textwidth}{!}{
    \begin{tabular}{ccccccccc}
        \hline
         & \multicolumn{2}{c}{Configuration} & & \multicolumn{5}{c}{PSNR/SSIM} \\
         & \textbf{Branches} & \textbf{Sample Size} & \textbf{Storage Size} & \textbf{Set5} & \textbf{Set14} & \textbf{BSDS100} & \textbf{Urban100} & \textbf{Manga109} \\
         \hline
         \ding{172}
         & \multirow{1}{*}{2} 
         & 3 & 2.710MB & 30.73/0.8677 & 27.68/0.7566 & 26.92/0.7130 & 24.54/0.7233 & 28.14/0.8680 \\
         \hline
         \ding{173} & \multirow{3}{*}{3} 
         & 3 & 4.065MB & 30.79/0.8693 & 27.72/0.7579 & 26.94/0.7142 & 24.57/0.7249 & 28.20/0.8696 \\
         \ding{174} & & 5 & 4.067MB & 30.85/0.8699 & 27.77/0.7584 & 26.96/0.7144 & 24.60/0.7257 & 28.27/0.8706 \\
         \ding{175} & & 7 & 4.069MB & 30.82/0.8694 & 27.72/0.7578 & 26.95/0.7146 & 24.58/0.7247 & 28.20/0.8693 \\
         \hline
         \ding{176} & \multirow{1}{*}{4}
         & 3 & 5.424MB & 30.83/0.8700 & 27.73/0.7586 & 26.96/0.7148 & 24.60/0.7262 & 28.26/0.8708 \\
         \hline

    \end{tabular}
    }
    \label{table:mulut-config-perf}
\end{table*}

\textbf{Qualitative Comparison.} As shown in Fig.~\ref{vis}, MuLUT+Ours produces clearer results, particularly on images of butterfly and the letter `e', where MuLUT often displays severe artifacts. In comparison with SPF-LUT+DFC, our SPF-Light achieves similar performance, while requiring less than half the storage space. 
\subsection{Ablation Study}

We conducted ablation experiments across different components and hyperparameters to investigate the impact of factors such as the number of branches, sample size, and the inclusion of AutoSample and AdaRL on performance.

\textbf{Sample Size.} We conducted experiments using different sample sizes 3, 5 and 7 with 3 branches, as shown in rows \ding{173}, \ding{174}, \ding{175} of the Table.~\ref{table:mulut-config-perf}. The sample size 5 achieved the empirically better results in terms of both PSNR and SSIM across all datasets. Additionally, as the sample size increased, storage requirements only increased marginally 0.002 MB per increment in sample size, demonstrating the method’s efficiency in handling larger receptive fields with negligible additional storage. However, when the sample size increased to 7, performance decreased slightly, with PSNR dropping by 0.03 dB on Set5. This may be due to the excessively large receptive field introducing irrelevant information that did not contribute effectively to the model's performance.

\textbf{Branches.} We conducted experiments using branches 2, 3 and 4 with sample size 3, as shown in rows \ding{172}, \ding{173}, \ding{176} of the Table.~\ref{table:mulut-config-perf}. As the number of branches increases, the PSNR progressively improves, reaching 30.73 dB, 30.79 dB and 30.83 dB on Set5. This improvement highlights the effectiveness of increasing the number of branches, which enhances the model's ability to capture detailed information. However, increasing the number of branches also leads to higher storage requirements. To strike a balance between performance and efficiency, we chose to use 3 branches with sample size 5 for MuLUT, as it provided empirically better performance while keeping storage manageable.
\begin{table*}[ht]
\centering
\renewcommand\arraystretch{1.1}
\caption{Ablation study on AutoSample and AdaRL for $\times$4 super-resolution.}
\begin{tabular}{cccccccc}
\hline
& Automatic Sampling & AdaRL & Set5 & Set14 & BSDS100 & Urban100 & Manga109 \\
\hline
\ding{172} & - & - & 30.60/0.8653 & 27.60/0.7541 & 26.86/0.7110 & 24.46/0.7194 & 27.90/0.8633 \\
\hline
\ding{173} & $\checkmark$ & - & 30.63/0.8657 & 27.63/0.7550 & 26.86/0.7112 & 24.49/0.7204 & 28.02/0.8627 \\
\hline
\ding{174} & - & $\checkmark$ & 30.70/0.8687 & 27.67/0.7572 & 26.89/0.7136 & 24.51/0.7242 & 28.07/0.8685 \\
\hline
\ding{175} & $\checkmark$ & $\checkmark$ & 30.79/0.8693 & 27.72/0.7579 & 26.94/0.7142 & 24.57/0.7249 & 28.20/0.8696 \\
\hline
\end{tabular}
\label{tab:ablation_auto_sample_adarl_set5}
\end{table*}

\textbf{AutoSample and AdaRL.} The ablation study results in Table.~\ref{tab:ablation_auto_sample_adarl_set5}, using 3 branches and sample size 3, highlight the individual and combined contributions of the AutoSample and AdaRL. For example, on Set5, adding only the AutoSample module results in a PSNR increase from 30.60 dB to 30.63 dB. Meanwhile, incorporating only the AdaRL module provides a more substantial boost, with PSNR increasing to 30.70 dB. When both modules are applied, performance reaches a PSNR of 30.79 dB. This demonstrates that each module enhances super-resolution independently, and their combination achieves the best results. The same trend is reflected across other datasets, reinforcing the complementary strengths of AutoSample and AdaRL.

\subsection{Inference Performance on Edge Device}

To systematically compare the performance of various super-resolution methods on mobile devices, we converted each model to ONNX format and tested them on an Android smartphone, specifically the OnePlus Ace 3 Pro running ColorOS 14. The super-resolution processing was performed on the Snapdragon 8 Gen 3 CPU’s Cortex-X4 core at 2.4 GHz, and the results were obtained when running $\times$4 super-resolution on a $224 \times 224$ image.

The experimental results are summarized in Table.~\ref{table:mobile-runtime}. For instance, MuLUT+Ours 1$\times$5 with 1 branch and sample size 5, reduces inference time significantly, from 5938.10 ms to 2043.95 ms, while also increasing PSNR by 0.02 dB compared to MuLUT. Similarly, SPF-Light reduces inference time from 31921.65 ms to 9910.95 ms, with a negligible PSNR decrease upon SPF-LUT+DFC. Notably, MuLUT+Ours achieves nearly the same inference time as MuLUT while increasing PSNR by 0.25 dB. Overall, our method achieves inference time reductions of about two-thirds on both MuLUT and SPF-LUT, with performance comparable to or even exceeding the original models.
These results are also visually represented in Fig.~\ref{fig:time_vs_perf}, our model demonstrates both lower inference time and improved performance. Our methods, represented in red, consistently occupy the Pareto optimal region in the upper-left quadrant, highlighting their advantage in achieving high efficiency while maintaining excellent performance.
\begin{table}[t]
  \centering
  \caption{Comparison of runtime and PSNR across different SR methods on Set5 on mobile device.}
\renewcommand\arraystretch{1.1}  
\setlength{\tabcolsep}{2mm}  
\resizebox{0.4\textwidth}{!}{%
    \begin{tabular}{ccccc}
        \hline
         &          \textbf{Runtime (ms)} & PSNR\\
         \hline
         MuLUT\cite{Li_2022_MuLUT} & 5938.1 & 30.60 \\
         MuLUT+Ours 1$\times$5 & 2043.95 & 30.62 \\
         MuLUT+Ours &5984.00 & 30.85 \\
         \hline
         SPF-LUT+DFC\cite{Li_2024_CVPR} &31921.65 & 31.05 \\
         SPF-Light (Ours) & 9910.95 & 31.02 \\
         \hline
    \end{tabular}
}
\label{table:mobile-runtime}
\end{table}

\definecolor{fig_blue}{RGB}{56, 84, 146}   
\definecolor{fig_red}{RGB}{212, 88, 88}    
\begin{figure}[t]
    \centering
    \includegraphics[width=0.8\columnwidth]{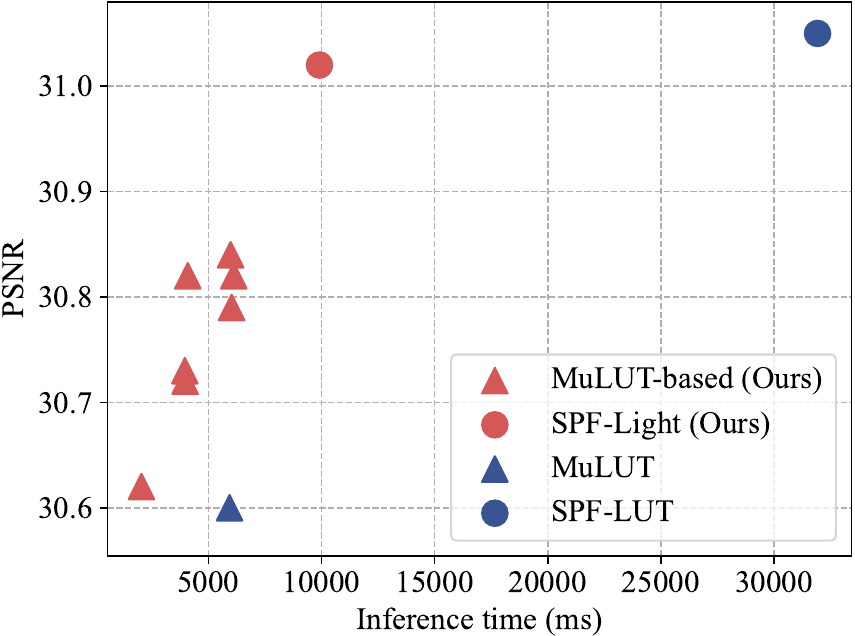}
    \caption{Comparison of inference time and performance of different SR methods.  Top-left is better. \textcolor{fig_red}{\textbf{Red}} denotes networks with our methods, and \textcolor{fig_blue}{\textbf{Blue}} means other methods. Our methods consistently occupy the Pareto optimal region in the upper-left quadrant, highlighting their advantage in achieving high efficiency while maintaining excellent performance.}
\label{fig:time_vs_perf}
\end{figure}
\section{Conclusion}
We propose two novel plug-and-play modules, AutoSample and AdaRL, to improve the performance of LUT-based super-resolution networks. AutoSample dynamically adapts the pixel extraction process by learning sampling weights during training, enabling it to expand the receptive field without increasing storage requirements and capture more detailed information. Meanwhile, AdaRL enhances the flow of information between layers through residual learning. Together, these modules enable significant reductions in storage space and inference time, making our approach particularly well-suited for deployment on resource-limited edge devices. The modular architecture of AutoSample and AdaRL allows for easy integration into a variety of LUT-based super-resolution networks, enabling efficient deployment of super-resolution models. Through experiments, we demonstrate the advantages of our method in terms of reduced storage size, improved performance, and faster runtime on resource-limited edge devices.

\section*{Acknowledgments}
 This work was supported by 
the National Natural Science Foundation of China (Grant No. 62402211) and the Natural Science Foundation of Jiangsu Province (Grant No. BK20241248). The authors would also like to thank the support from the Collaborative Innovation Center of Novel Software Technology and Industrialization, Jiangsu, China.

{
    \small
    \bibliographystyle{ieeenat_fullname}
    \bibliography{main}

\begin{thebibliography}{42}
\providecommand{\natexlab}[1]{#1}
\providecommand{\url}[1]{\texttt{#1}}
\expandafter\ifx\csname urlstyle\endcsname\relax
  \providecommand{\doi}[1]{doi: #1}\else
  \providecommand{\doi}{doi: \begingroup \urlstyle{rm}\Url}\fi

\bibitem[Agustsson and Timofte(2017)]{Agustsson_2017_CVPR_Workshops}
Eirikur Agustsson and Radu Timofte.
\newblock Ntire 2017 challenge on single image super-resolution: Dataset and study.
\newblock In \emph{The IEEE Conference on Computer Vision and Pattern Recognition (CVPR) Workshops}, 2017.

\bibitem[Ahn et~al.(2018)Ahn, Kang, and Sohn]{DBLP:conf/eccv/AhnKS18}
Namhyuk Ahn, Byungkon Kang, and Kyung{-}Ah Sohn.
\newblock Fast, accurate, and lightweight super-resolution with cascading residual network.
\newblock In \emph{ECCV}, 2018.

\bibitem[Bertsekas et~al.(2003)Bertsekas, Nedic, and Ozdaglar]{bertsekas2003convex}
Dimitri Bertsekas, Angelia Nedic, and Asuman Ozdaglar.
\newblock \emph{Convex analysis and optimization}.
\newblock Athena Scientific, 2003.

\bibitem[Bevilacqua et~al.(2012)Bevilacqua, Roumy, Guillemot, and Alberi{-}Morel]{DBLP:conf/bmvc/BevilacquaRGA12}
Marco Bevilacqua, Aline Roumy, Christine Guillemot, and Marie{-}Line Alberi{-}Morel.
\newblock Low-complexity single-image super-resolution based on nonnegative neighbor embedding.
\newblock In \emph{BMVC}, 2012.

\bibitem[Chang et~al.(2004)Chang, Yeung, and Xiong]{DBLP:conf/cvpr/ChangYX04}
Hong Chang, Dit{-}Yan Yeung, and Yimin Xiong.
\newblock Super-resolution through neighbor embedding.
\newblock In \emph{CVPR}, 2004.

\bibitem[Chen et~al.(2019)Chen, Xiong, Tian, Zha, and Wu]{DBLP:conf/cvpr/ChenXTZW19}
Chang Chen, Zhiwei Xiong, Xinmei Tian, Zheng{-}Jun Zha, and Feng Wu.
\newblock Camera lens super-resolution.
\newblock In \emph{CVPR}, 2019.

\bibitem[Cheng et~al.(2021)Cheng, Xiong, Chen, Liu, and Zha]{DBLP:conf/cvpr/ChengXC0Z21}
Zhen Cheng, Zhiwei Xiong, Chang Chen, Dong Liu, and Zheng{-}Jun Zha.
\newblock Light field super-resolution with zero-shot learning.
\newblock In \emph{CVPR}, 2021.

\bibitem[Dong et~al.(2014)Dong, Loy, He, and Tang]{DBLP:conf/eccv/DongLHT14}
Chao Dong, Chen~Change Loy, Kaiming He, and Xiaoou Tang.
\newblock Learning a deep convolutional network for image super-resolution.
\newblock In \emph{ECCV}, 2014.

\bibitem[Dong et~al.(2016{\natexlab{a}})Dong, Loy, He, and Tang]{DBLP:journals/pami/DongLHT16}
Chao Dong, Chen~Change Loy, Kaiming He, and Xiaoou Tang.
\newblock Image super-resolution using deep convolutional networks.
\newblock \emph{{IEEE} Trans. Pattern Anal. Mach. Intell.}, 38\penalty0 (2):\penalty0 295--307, 2016{\natexlab{a}}.

\bibitem[Dong et~al.(2016{\natexlab{b}})Dong, Loy, and Tang]{DBLP:conf/eccv/DongLT16}
Chao Dong, Chen~Change Loy, and Xiaoou Tang.
\newblock Accelerating the super-resolution convolutional neural network.
\newblock In \emph{ECCV}, 2016{\natexlab{b}}.

\bibitem[Freeman et~al.(2000)Freeman, Pasztor, and Carmichael]{DBLP:journals/ijcv/FreemanPC00}
William~T. Freeman, Egon~C. Pasztor, and Owen~T. Carmichael.
\newblock Learning low-level vision.
\newblock \emph{Int. J. Comput. Vis.}, 40\penalty0 (1):\penalty0 25--47, 2000.

\bibitem[Freeman et~al.(2002)Freeman, Jones, and Pasztor]{DBLP:journals/cga/FreemanJP02}
William~T. Freeman, Thouis~R. Jones, and Egon~C. Pasztor.
\newblock Example-based super-resolution.
\newblock \emph{{IEEE} Computer Graphics and Applications}, 22\penalty0 (2):\penalty0 56--65, 2002.

\bibitem[He et~al.(2016)He, Zhang, Ren, and Sun]{DBLP:conf/cvpr/HeZRS16}
Kaiming He, Xiangyu Zhang, Shaoqing Ren, and Jian Sun.
\newblock Deep residual learning for image recognition.
\newblock In \emph{2016 {IEEE} Conference on Computer Vision and Pattern Recognition, {CVPR} 2016, Las Vegas, NV, USA, June 27-30, 2016}, pages 770--778. {IEEE} Computer Society, 2016.

\bibitem[Ho et~al.(2020)Ho, Jain, and Abbeel]{DBLP:conf/nips/HoJA20}
Jonathan Ho, Ajay Jain, and Pieter Abbeel.
\newblock Denoising diffusion probabilistic models.
\newblock In \emph{Advances in Neural Information Processing Systems 33: Annual Conference on Neural Information Processing Systems 2020, NeurIPS 2020, December 6-12, 2020, virtual}, 2020.

\bibitem[Huang et~al.(2015)Huang, Singh, and Ahuja]{DBLP:conf/cvpr/HuangSA15}
Jia{-}Bin Huang, Abhishek Singh, and Narendra Ahuja.
\newblock Single image super-resolution from transformed self-exemplars.
\newblock In \emph{CVPR}, 2015.

\bibitem[Jo and Kim(2021)]{DBLP:conf/cvpr/JoK21}
Younghyun Jo and Seon~Joo Kim.
\newblock Practical single-image super-resolution using look-up table.
\newblock In \emph{CVPR}, 2021.

\bibitem[Keys(1981)]{Keys1981CubicCI}
R. Keys.
\newblock Cubic convolution interpolation for digital image processing.
\newblock \emph{IEEE Transactions on Acoustics, Speech, and Signal Processing}, 29:\penalty0 1153--1160, 1981.

\bibitem[Kim et~al.(2016)Kim, Lee, and Lee]{DBLP:conf/cvpr/KimLL16a}
Jiwon Kim, Jung~Kwon Lee, and Kyoung~Mu Lee.
\newblock Accurate image super-resolution using very deep convolutional networks.
\newblock In \emph{CVPR}, 2016.

\bibitem[Li et~al.(2022)Li, Chen, Cheng, and Xiong]{Li_2022_MuLUT}
Jiacheng Li, Chang Chen, Zhen Cheng, and Zhiwei Xiong.
\newblock {MuLUT}: Cooperating multiple look-up tables for efficient image super-resolution.
\newblock In \emph{ECCV}, 2022.

\bibitem[Li et~al.(2024)Li, Li, and Xiong]{Li_2024_CVPR}
Yinglong Li, Jiacheng Li, and Zhiwei Xiong.
\newblock Look-up table compression for efficient image restoration.
\newblock In \emph{Proceedings of the IEEE/CVF Conference on Computer Vision and Pattern Recognition (CVPR)}, pages 26016--26025, 2024.

\bibitem[Lim et~al.(2017)Lim, Son, Kim, Nah, and Lee]{DBLP:conf/cvpr/LimSKNL17}
Bee Lim, Sanghyun Son, Heewon Kim, Seungjun Nah, and Kyoung~Mu Lee.
\newblock Enhanced deep residual networks for single image super-resolution.
\newblock In \emph{CVPRW}, 2017.

\bibitem[Ma et~al.(2022)Ma, Zhang, Zhou, and Lu]{DBLP:conf/eccv/MaZZL22}
Cheng Ma, Jingyi Zhang, Jie Zhou, and Jiwen Lu.
\newblock Learning series-parallel lookup tables for efficient image super-resolution.
\newblock In \emph{Computer Vision - {ECCV} 2022 - 17th European Conference, Tel Aviv, Israel, October 23-27, 2022, Proceedings, Part {XVII}}, pages 305--321. Springer, 2022.

\bibitem[Martin et~al.(2001)Martin, Fowlkes, Tal, and Malik]{DBLP:conf/iccv/MartinFTM01}
David~R. Martin, Charless~C. Fowlkes, Doron Tal, and Jitendra Malik.
\newblock A database of human segmented natural images and its application to evaluating segmentation algorithms and measuring ecological statistics.
\newblock In \emph{ICCV}, 2001.

\bibitem[Matsui et~al.(2017)Matsui, Ito, Aramaki, Fujimoto, Ogawa, Yamasaki, and Aizawa]{DBLP:journals/mta/MatsuiIAFOYA17}
Yusuke Matsui, Kota Ito, Yuji Aramaki, Azuma Fujimoto, Toru Ogawa, Toshihiko Yamasaki, and Kiyoharu Aizawa.
\newblock Sketch-based manga retrieval using manga109 dataset.
\newblock \emph{Multim. Tools Appl.}, 76\penalty0 (20):\penalty0 21811--21838, 2017.

\bibitem[Qiu(2000)]{DBLP:journals/jvcir/Qiu00}
Guoping Qiu.
\newblock Interresolution look-up table for improved spatial magnification of image.
\newblock \emph{J. Vis. Commun. Image Represent.}, 11\penalty0 (4):\penalty0 360--373, 2000.

\bibitem[Romano et~al.(2017)Romano, Isidoro, and Milanfar]{DBLP:journals/tci/RomanoIM17}
Yaniv Romano, John Isidoro, and Peyman Milanfar.
\newblock {RAISR:} rapid and accurate image super resolution.
\newblock \emph{{IEEE} Trans. Computational Imaging}, 3\penalty0 (1):\penalty0 110--125, 2017.

\bibitem[Song et~al.(2018)Song, Xiong, Liu, Xiong, Wu, and Gao]{DBLP:journals/tip/SongXLXWG18}
Qiang Song, Ruiqin Xiong, Dong Liu, Zhiwei Xiong, Feng Wu, and Wen Gao.
\newblock Fast image super-resolution via local adaptive gradient field sharpening transform.
\newblock \emph{IEEE TIP}, 27\penalty0 (4):\penalty0 1966--1980, 2018.

\bibitem[Timofte et~al.(2013)Timofte, Smet, and Gool]{DBLP:conf/iccv/TimofteDG13}
Radu Timofte, Vincent~De Smet, and Luc~Van Gool.
\newblock Anchored neighborhood regression for fast example-based super-resolution.
\newblock In \emph{ICCV}, 2013.

\bibitem[Timofte et~al.(2014)Timofte, Smet, and Gool]{DBLP:conf/accv/TimofteSG14}
Radu Timofte, Vincent~De Smet, and Luc~Van Gool.
\newblock {A+:} adjusted anchored neighborhood regression for fast super-resolution.
\newblock In \emph{ACCV}, 2014.

\bibitem[Wang et~al.(2014)Wang, Wu, and Pan]{DBLP:journals/tip/WangWP14}
LingFeng Wang, Huai{-}Yu Wu, and Chunhong Pan.
\newblock Fast image upsampling via the displacement field.
\newblock \emph{IEEE TIP}, 23\penalty0 (12):\penalty0 5123--5135, 2014.

\bibitem[Wang et~al.(2018)Wang, Yu, Wu, Gu, Liu, Dong, Qiao, and Loy]{DBLP:conf/eccv/WangYWGLDQL18}
Xintao Wang, Ke Yu, Shixiang Wu, Jinjin Gu, Yihao Liu, Chao Dong, Yu Qiao, and Chen~Change Loy.
\newblock {ESRGAN:} enhanced super-resolution generative adversarial networks.
\newblock In \emph{ECCV Workshops}, 2018.

\bibitem[Xiao et~al.(2020)Xiao, Xiong, Fu, Liu, and Zha]{DBLP:conf/mm/XiaoXFLZ20}
Zeyu Xiao, Zhiwei Xiong, Xueyang Fu, Dong Liu, and Zheng{-}Jun Zha.
\newblock Space-time video super-resolution using temporal profiles.
\newblock In \emph{{ACM} MM}, 2020.

\bibitem[Xiao et~al.(2021)Xiao, Fu, Huang, Cheng, and Xiong]{DBLP:conf/cvpr/XiaoFHCX21}
Zeyu Xiao, Xueyang Fu, Jie Huang, Zhen Cheng, and Zhiwei Xiong.
\newblock Space-time distillation for video super-resolution.
\newblock In \emph{CVPR}, 2021.

\bibitem[Xiong et~al.(2010)Xiong, Sun, and Wu]{DBLP:journals/tip/XiongSW10a}
Zhiwei Xiong, Xiaoyan Sun, and Feng Wu.
\newblock Robust web image/video super-resolution.
\newblock \emph{IEEE TIP}, 19\penalty0 (8):\penalty0 2017--2028, 2010.

\bibitem[Xiong et~al.(2013)Xiong, Xu, Sun, and Wu]{DBLP:journals/tmm/XiongXSW13}
Zhiwei Xiong, Dong Xu, Xiaoyan Sun, and Feng Wu.
\newblock Example-based super-resolution with soft information and decision.
\newblock \emph{{IEEE} Trans. Multim.}, 15\penalty0 (6):\penalty0 1458--1465, 2013.

\bibitem[Xu et~al.(2021)Xu, Xiao, Yao, Zhang, and Xiong]{DBLP:conf/mm/XuXYZX21}
Ruikang Xu, Zeyu Xiao, Mingde Yao, Yueyi Zhang, and Zhiwei Xiong.
\newblock Stereo video super-resolution via exploiting view-temporal correlations.
\newblock In \emph{{ACM} MM}, 2021.

\bibitem[Yang et~al.(2010)Yang, Wright, Huang, and Ma]{DBLP:journals/tip/YangWHM10}
Jianchao Yang, John Wright, Thomas~S. Huang, and Yi Ma.
\newblock Image super-resolution via sparse representation.
\newblock \emph{IEEE TIP}, 19\penalty0 (11):\penalty0 2861--2873, 2010.

\bibitem[Zeyde et~al.(2010{\natexlab{a}})Zeyde, Elad, and Protter]{DBLP:conf/cas/ZeydeEP10}
Roman Zeyde, Michael Elad, and Matan Protter.
\newblock On single image scale-up using sparse-representations.
\newblock In \emph{Curves and Surfaces}, 2010{\natexlab{a}}.

\bibitem[Zeyde et~al.(2010{\natexlab{b}})Zeyde, Elad, and Protter]{zeyde2010single}
Roman Zeyde, Michael Elad, and Matan Protter.
\newblock On single image scale-up using sparse-representations.
\newblock In \emph{International conference on curves and surfaces}, pages 711--730. Springer, 2010{\natexlab{b}}.

\bibitem[Zhang et~al.(2019)Zhang, Liu, and Xiong]{DBLP:conf/iccv/ZhangLX19}
Haochen Zhang, Dong Liu, and Zhiwei Xiong.
\newblock Two-stream action recognition-oriented video super-resolution.
\newblock In \emph{ICCV}, 2019.

\bibitem[Zhang et~al.(2018{\natexlab{a}})Zhang, Li, Li, Wang, Zhong, and Fu]{DBLP:conf/eccv/ZhangLLWZF18}
Yulun Zhang, Kunpeng Li, Kai Li, Lichen Wang, Bineng Zhong, and Yun Fu.
\newblock Image super-resolution using very deep residual channel attention networks.
\newblock In \emph{ECCV}, 2018{\natexlab{a}}.

\bibitem[Zhang et~al.(2018{\natexlab{b}})Zhang, Tian, Kong, Zhong, and Fu]{DBLP:conf/cvpr/ZhangTKZ018}
Yulun Zhang, Yapeng Tian, Yu Kong, Bineng Zhong, and Yun Fu.
\newblock Residual dense network for image super-resolution.
\newblock In \emph{CVPR}, 2018{\natexlab{b}}.

\end{thebibliography}
}


\end{document}